\documentclass[10pt, conference, compsocconf]{IEEEtran}
\usepackage{cite}

\ifCLASSINFOpdf
   \usepackage[pdftex]{graphicx}
\else
   \usepackage[dvips]{graphicx}
\fi
\usepackage[cmex10]{amsmath}
\usepackage{amssymb}

\usepackage[font=footnotesize]{caption}
\usepackage[font=footnotesize]{subfig}

 \usepackage{url}

\usepackage{multirow}
\usepackage{booktabs}

\begin{document}
%
\title{Human Motion Prediction via Pattern Completion in Latent Representation Space}

\author{\IEEEauthorblockN{Yi Tian Xu$^{1,2}$, 
Yaqiao Li$^2$ and
David Meger$^{1,2}$}
 \IEEEauthorblockA{
$^1$Mobile Robotics Lab\\
$^2$School of Computer Science\\
McGill University\\
Montreal, Canada\\
\{yi.t.xu, yaqiao.li\}@mail.mcgill.ca, dmeger@cim.mcgill.ca}
}

\newcommand\blfootnote[1]{%
  \begingroup
  \renewcommand\thefootnote{}\footnote{#1}%
  \addtocounter{footnote}{-1}%
  \endgroup
}

\maketitle

\begin{abstract}
Inspired by ideas in cognitive science, we propose a novel and general approach to solve human motion understanding via pattern completion on a learned latent representation space. Our model outperforms current state-of-the-art methods in human motion prediction across a number of tasks, with no customization. To construct a latent representation for time-series of various lengths, we propose a new and generic autoencoder based on sequence-to-sequence learning. While traditional inference strategies find a correlation between an input and an output, we use pattern completion, which views the input as a partial pattern and to predict the best corresponding complete pattern. Our results demonstrate that this approach has advantages when combined with our autoencoder in solving human motion prediction, motion generation and action classification.
\blfootnote{\textcopyright 2019 IEEE. Personal use of this material is permitted. Permission from IEEE must be obtained for all other uses, in any current or future media, including reprinting/republishing this material for advertising or promotional purposes,creating new collective works, for resale or redistribution to servers or lists, or reuse of any copyrighted component of this work in other works.}
\end{abstract}


%
\IEEEpeerreviewmaketitle

\section{Introduction}

Knowledge of how humans move can help intelligent robots in tasks involving an interactive human environment, such as navigating through a crowded street or playing sports and tabletop games with humans.  
Capturing human motion requires feature detection and tracking, as well as modeling a complex dynamical structure which is highly
non-linear, spontaneous and entangled with physical constraints, intention and high-level semantics. With the arrival of large motion capture databases, such as the Human3.6M dataset \cite{ionescu2014human3}, and 3D pose estimation algorithms, research has focused on the core patterns present in human motion rather than the distracting visual features. Recently, a series of skeleton-based deep learning methods have greatly increased the performance for human motion prediction, while introducing increasingly specialized and sophisticated model designs. 

We attack skeleton-based human motion prediction using a \emph{general} representation learning approach \cite{bengio2013representation} without relying on specialized architectures or external knowledge on the data structure. Our method consists of two coordinated steps. First, we learn a latent representation space using a hierarchical sequence-to-sequence (Seq2Seq) architecture, to reveal the underlying structure in the complex Human3.6M dataset. Then, we use the learned representations for motion prediction and for two related tasks: motion generation and action classification, through a process called \emph{pattern completion} \cite{barsalou2005situated}. The latter is the core idea in our method; instead of viewing inference as finding the correlation between an input and an output, we view the input as a partial pattern that is to be completed.

\begin{figure}[t] 
\centering
    \includegraphics[width=\linewidth]{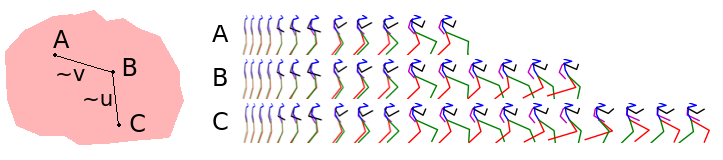}
    \caption{By encoding motion sequences into a well-structured latent space, we are able to complete the pattern $A$ to a pattern that approximates $B$ by simply using vector addition, e.g. $A+v\approx B$, where $v$ is a vector that can be directly computed. The process is recursive and can be repeated to extend to motion $C$.}  \label{fig:embedding} 
\end{figure}

As proposed in situated conceptualization in cognitive science \cite{barsalou2005situated},
pattern completion can support diverse forms of intelligent tasks and provides an important grounding of new situations into experienced situations, rendering structure to the latent representations. To our knowledge, we are the first to attempt this conceptual contribution in the domain of human motion understanding. See an illustration of pattern completion in Fig. \ref{fig:embedding}.

In summary, our main contributions are:
\begin{enumerate}
    \item We propose a new and generic autoencoder that uses a hierarchical Seq2Seq structure to construct latent representations of time-series of various lengths while maintaining a well-structured embedding. 

    \item We implement pattern completion on the latent representation space with either a single-layer network or vector addition for human motion prediction, motion generation and action classification. We often achieve high performance. 
\end{enumerate}

Our results show competitive and sometimes higher performance compared to state-of-the-art methods in human motion prediction. In particular, our method outperforms the state of the arts for some aperiodic actions in Human3.6M such as \emph{greeting}, \emph{sitting} and \emph{taking photo}, which are notoriously challenging. Furthermore, the representations learned by our method also allow for the motion generation and action classification tasks to be performed effectively.

\section{Related works}   \label{sec:RelatedWorks}

\subsection{Human motion prediction with Human3.6M}    \label{sec:hmp}

The Human3.6M dataset \cite{ionescu2014human3} is one the largest and most challenging benchmarks to evaluate human motion understanding. Its large variety of poses are recorded from 7 professional actors doing 15 activities, including walking, eating, smoking and engaging in a discussion. Due to the stochastic nature of human movement, previous authors have separated motion prediction into two sub-tasks: short-term and long-term prediction. Short-term prediction is commonly compared quantitatively with mean angle error, while long-term predictions are usually assessed qualitatively. This is because even human intelligence is not able to uniquely determine the motion of a character tens of seconds into the future but rather only capture a sampling of plausible outcomes. 


Recurrent neural networks (RNN) are commonly used to solve both short-term and long-term prediction. Earlier works \cite{fragkiadaki2015recurrent, jain2016structural} suffer from a noticeable discontinuity between the end of the input (last observed frame) and the first predicted frame.
Martinez et al. \cite{martinez2017human} alleviates this problem using a Seq2Seq model with a residual connection, boosting the performance for short-term prediction. 

Recent works focus on long-term prediction which often collapses to an undesired common pose, especially for aperiodic motions.
This failure is possibly caused by the common use of mean squared error (MSE), a high-tailed loss that discourages making risky predictions \cite{martinez2017human}. Additionally, MSE, as well as other traditional losses, treat each joint with equal weight. In reality, the impact of joints on any given motion is non-uniform. This motivates many complex or data-specialized loss function. 
Pavllo et al. \cite{pavllo2018quaternet} propose to perform Forward Kinematics during training to compute the loss in Cartesian space.
Gui et al. \cite{gui2018adversarial} introducing geodesic loss to capture the geometric structure of the angles.
Several authors \cite{lin2018human, pavllo2018quaternet, li2018convolutional} also propose to tackle the short-term and long-term prediction tasks separately, or to have two model variations, each adapted and optimized for one of the tasks. 

Furthermore, RNNs of all flavors, including both Long Short-Term Memory (LSTM) and Gated Recurrent Units (GRU), are hypothesized to have difficulty in keeping track of long-term information \cite{tang2018long} and spatial correlation (e.g. between left and right arms) \cite{li2018convolutional}. Tang et al. \cite{tang2018long} approach this problem by adding an attention unit and their Modified Highway Unit (MHU) to summarize motion history and to focus on joints with large motions. Li et al. \cite{li2018convolutional} use convolutional filters to learn spatio-temporal dependencies.  

In our work, we design a single model to address both short-term and long-term prediction without additional data-specialized or task-specialized architectures. We also employ a Hierarchical Seq2Seq architecture \cite{li2015hierarchical} to avoid the loss of long-term information.


\subsection{Representation learning methods}   \label{sec:representation}

Several deep learning methods include objectives that encourage reproduction of the input data, including autoencoders, Variational Autoencoders (VAE) \cite{doersch2016tutorial} and Generative Adversarial Networks (GAN) \cite{goodfellow2014generative}. These approaches are known to produce a latent space that has certain meaningful structures. Nearby latent representations are similar in the original data representation space. Thus, interpolation on the latent space usually produces smooth transitions on the original space, and clustering on the latent space often produces semantically meaningful groups. In particular, successful distributional semantics models in NLP \cite{mikolov2013efficient, mikolov2013distributed, pennington2014glove} demonstrate the additive compositionality property, enabling the word analogy task to be solvable using vector addition on the latent space.

Previous works in human motion representation learning such as \cite{holden2015learning, butepage2017deep} show how the learned representation can be used for various tasks such as fixing corrupted data or performing action classification. However, to our knowledge, none of them demonstrate performance comparable to the current state of the art in human motion prediction.

\section{Method}    \label{sec:ourMethod}

\begin{figure*}[h]  
\centering
\subfloat[H-Seq2SeqsAE architecture.]
{
    \includegraphics[width=0.72\textwidth]{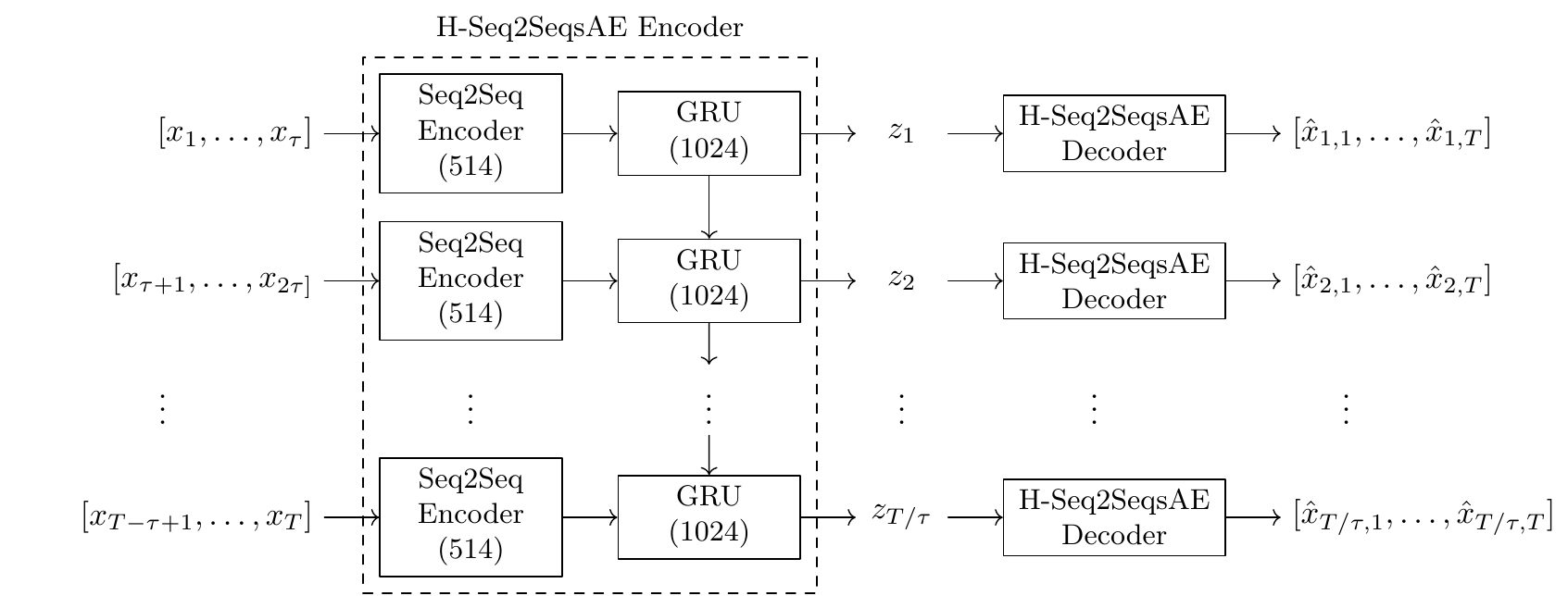}
    \label{fig:arch}
}
\subfloat[Details for the decoder.]
{
    \includegraphics[width=0.26\textwidth]{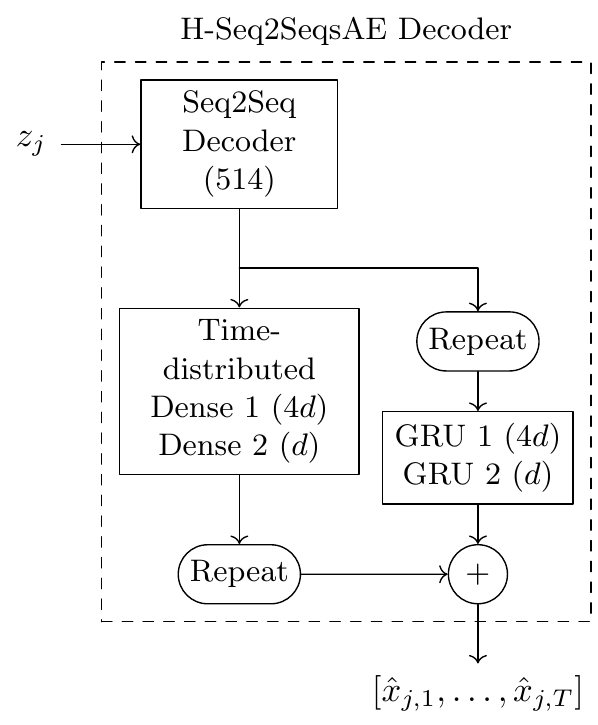}
    \label{fig:decoder}
}
 \caption{An illustration of our Hierarchical Sequence-to-Sequence Autoencoder (H-Seq2SeqsAE). We use common Seq2Seq encoder and decoder \cite{sutskever2014sequence} in our model. The Repeat unit in (b) takes input a sequence of length $T/\tau$ and outputs a sequence of length $T$, where each element from the input is repeated consecutively $\tau$ times. The output dimensionalities are specified in parenthesis. When training for $T=60$, we use 1500 dimensions for the gated recurrent unit (GRU) in (a). In (b), $d$ denotes the number of features. Note that in (a), the weights are shared across rows. 
}
 \label{fig:model}
\end{figure*}

Our method consists of two steps: representation learning and pattern completion. In the representation learning step, we learn encoding and decoding functions $E$ and $D$, which map observations to corresponding latent representations, and back. In our novel pattern completion step, we train a pattern completion function $G$ to map inputs to their most sensible full patterns in the well-structured latent space. To make overall predictions, we apply the function $D\circ G \circ E$ to encode, complete and decode.  Specifically, given an input data pair $(X,Y)$, the function $G$ aims to predict the latent representation of $XY$ from the latent representation of $X$. This is in contrast to the common practice that predicts $Y$ directly from $X$. Below, we explain the details of our method.

Let $S = \{(X_i,Y_i)\}_{i=1}^N$ be a given dataset of input and ground truth pairs. For the motion prediction task, $X_i$ is the beginning of a motion that is observed in order to complete the unseen portion $Y_i$. Each $X_i$ and each $Y_i$ are represented as a sequence of joint angles. We wish to learn a prediction function $F$ such that
\begin{equation}
    \{X_i\}_{i=1}^N \xrightarrow{F} \{\widehat{Y}_i\}_{i=1}^N
\end{equation}
where the difference between $Y_i$ and $\widehat{Y}_i$ is minimized.

In the representation learning step, we construct $S'= \{X_i, X_iY_i\}_{i=1}^N$ where $X_iY_i$ denotes $X_i$ concatenated with $Y_i$. We learn an encoding and a decoding function, $E$ and $D$, respectively, such that $E$ maps every element in $S'$ to a latent representation in some lower-dimensional space $Z\subseteq \mathbb{R}^n$,
\begin{equation}   \label{eq:S1}
    S' \xrightarrow{E} Z \xrightarrow{D} \widehat{S'}.
\end{equation}

In the pattern completion step, we view $E(X_i)$ as a partial pattern and $E(X_iY_i)$ as a complete pattern, respectively, and use the pre-trained $E$ and $D$ to learn a \emph{pattern completion} function $G$ on the representation space. Specifically, let $P = \{p_i = E(X_i)\}_{i=1}^N$ be the set of partial patterns, and $C = \{c_i = E(X_iY_i)\}_{i=1}^N$ be the set of complete patterns. Note that $P, C \subseteq Z$. We learn a function $G$ such that
\begin{equation}   \label{eq:G}
    P \xrightarrow{G} \widehat{C},
\end{equation}
where the difference between $c_i$ and $\widehat{c_i}$ is minimized. That is, we wish to predict the complete pattern. Finally we obtain $F = D\circ G \circ E$. That is,
\begin{equation}   \label{eq:S2}
    \{X_i\}_{i=1}^N \xrightarrow{E} P \xrightarrow{G}  \widehat{C}  \xrightarrow{D} \{\widehat{X}_i\widehat{Y}_i\} \subseteq \widehat{S'}.
\end{equation}

Note that this step reflects our choice to map $X_i$ to $\widehat{X}_i\widehat{Y}_i$ as suggested by pattern completion in cognitive science \cite{barsalou2005situated}, rather than to map $X_i$ to $\widehat{Y}_i$ directly as in standard learning methods. 
While many tasks only require $\widehat{Y}_i$, we show that 
training for completion yields improved performance (see Section \ref{sec:ablation}).
To distinguish between completing $X$ with $XY$ and matching $X$ to $Y$, we call the latter \emph{pattern matching}.

To this point, our discussion has treated the input motion data as simple vectors, but it is crucial to capture their time series nature. In order for $E$ to encode sequences of different lengths, we modify the standard autoencoding framework so that subsequences of the input sequence can be mapped to latent representations. Section \ref{sec:StoS} presents the details of our autoencoder.

One crucial assumption in our method is that after learning $E$ and $D$, $G$ can be modeled with a much simpler function than completing in raw space and can be learned in a short amount of time. In our experiments, we model $G$ using either a single dense layer network or vector addition. In Section \ref{sec:RepMatching-Predict}, we describe how $G$ can be implemented using vector addition.

\subsection{Hierarchical Sequence-to-Sequences Autoencoder (H-Seq2SeqsAE)}    \label{sec:StoS}
Our model takes a sequence as its input and outputs multiple sequences, each corresponding to a reconstruction of a subsequence from the original input. It is based on Hierarchical Seq2Seq model \cite{li2015hierarchical} in order to avoid losing long-term historical information \cite{tang2018long, li2018convolutional}. We name our autoencoder \emph{Hierarchical Sequence-to-Sequences Autoencoder (H-Seq2SeqsAE)}. 

More specifically, let $T$ be the length of an input sequence and $\tau$ be a divisor of $T$. Given a sequence of joint angles $X = [x_1, x_2, \dots, x_T]$, we partition $X$ into $T/\tau$ subsequences  $[x_1, \dots, x_{\tau}], \dots, [x_{T-\tau+1}, \dots, x_T]$.

For the encoder, our model first obtains sub-encodings for these subsequences using a standard Seq2Seq encoder 
\cite{sutskever2014sequence}. Next, the $T/\tau$ sub-encodings are fed into a higher-level encoder which outputs $T/\tau$ encodings such that
the $j$th encoding, $z_j$, corresponds to $[x_1, \dots, x_{j\tau}]$. 
See Fig. \ref{fig:arch}. 

For the decoder, we modify the Hierarchical Seq2Seq decoder \cite{li2015hierarchical} using residual connections, which were shown to improve the performance in \cite{martinez2017human, pavllo2018quaternet}. Given $z_j$, we first apply the standard Seq2Seq decoder to obtain a sequence of length $T/\tau$. Each element in the sequence is then passed to two dense layers to obtain a pose. Another pathway leads the entire sequence to two RNNs to obtain $T$ residuals. Finally, the decoder outputs the combined poses and residuals. 
See Fig. \ref{fig:decoder}.

Since an important part of our decoder is based on the residual angles, a natural output for our model when autoencoding a subsequence of length $j\tau < T$ is zero motion after the $j\tau^{th}$ frame. Therefore, we use the following loss function
\begin{equation}
    \frac{1}{T/\tau} \sum_{j=1}^{T/\tau} l([x_1, x_2, \dots, x_{j\tau}, x_{j\tau}, \dots, x_{j\tau}], [\hat{x}_{j,1}, \dots, \hat{x}_{j,T}]),
\end{equation}
in which $[x_1, x_2, \dots, x_{j\tau}, x_{j\tau}, \dots, x_{j\tau}]$ is also a sequence of length $T$. We encode the moment when a motion stops rather than the exact length of the sequence. We use mean absolute error (MAE) for $l$ as it does not pose specific assumptions or constraints on the data format, and we find that it performs better compared to MSE in our method.

Finally, we define our functions $E$ and $D$ as follows. Given input sequence $X$ of length $j\tau \leq T$, we construct $X'$ by appending $X$ with placeholders to reach length $T$. We feed $X'$ to our H-Seq2SeqsAE encoder and take the $j^{th}$ output from it as the output for $E$. Note this allows $E$ to take input of various lengths: $\tau, 2\tau, \ldots, j\tau, \ldots, T$.
For $D$, we define it as the H-Seq2SeqsAE decoder.

\subsection{Pattern completion using vector addition}    \label{sec:RepMatching-Predict}


The emerging structure in the latent representation space allows for simple and intuitive vector addition to accurately predict human motion. See Fig. \ref{fig:embedding} for an illustration of this operation at work.

Given a set of input and ground truth pairs $S_j = \{(X,Y): |X| = j\tau, |XY| = T\}$ for some $j$. We define
\begin{equation}
    d_{j}(X,Y) = E(XY) - E(X).
\end{equation}
For an arbitrary input sequence $X'$ such that $|X'| = j\tau$ and $|X'Y'| = T$, we simply use the following vector addition
\begin{equation}
    E(X') + v_j
\end{equation}
to approximate $E(X'Y')$, where 
\begin{equation}
    v_j = \frac{1}{|S_j|}\sum_{(X,Y)\in S_j} d_{j}(X,Y).
\end{equation}
In other words, $v_j$ is the average difference between the latent representations of all $X$ and $XY$ seen in $S_j$. As we observe, the variance of $d_{j}(X,Y)$ is low (see Section \ref{sec:ablation}), and each $v_j$ can be computed using a small sample (e.g. using 1000 samples as in Fig. \ref{fig:training}) to obtain high quality results. 

Such additive relationship between $X$ and $XY$ in our latent representation space is analogous to the additive compositionality defined by Mikolov et al. \cite{mikolov2013distributed}. As our H-Seq2SeqsAE captures the robust features of $X$ and $XY$, we find a stronger correlation between $E(X)$ and $E(XY)$ than between $E(X)$ and $E(Y)$ (as shown in Section \ref{sec:ablation}).



\subsection{Action classification and label recovery}

\label{sec:RepMatching-Class}

To include action label information, we concatenate a one-hot encoded action type vector with each pose, similar to recent literature \cite{martinez2017human, gopalakrishnan2018neural, gui2018adversarial}. With the action label and human motion learned by our autoencoder, this knowledge can be used to solve the action classification task. We apply our pattern completion method to action classification in two variations.


In the first variation, our H-Seq2SeqsAE learns to encode both the supervised and unsupervised motion sequences, and the action label itself. To achieve this, at each epoch of the training, we randomly choose a third of the data and set the label vector to zero. Another third is randomly chosen with the poses set to zeros. 

In the second variation, our H-Seq2SeqsAE learns to encode the supervised motion sequences. Classification by this variation is, therefore, more similar to fixing corrupted data or filling missing information, thus we call it label recovery.

\section{Experiments}              \label{sec:Exp}

\bgroup
\def\tabcolsep{3pt}
\begin{table*}[h]
\centering
\caption{Comparison of mean angle error between our method and top performing baselines for short-term motion prediction. The ``Average*'' column is the average error over all 15 actions.} \label{Table:result40}
\subfloat[Short-term prediction with 30 input frames and normalized angles between -1 and 1.]
{
    \resizebox{\textwidth}{!}{
        
\begin{tabular}{l|c c c c|c c c c|c c c c|c c c c||c c c c}
    \toprule
    & \multicolumn{4}{c|}{Walking} 
    & \multicolumn{4}{c|}{Eating} 
    & \multicolumn{4}{c|}{Smoking} 
    & \multicolumn{4}{c||}{Discussion} 
    & \multicolumn{4}{c}{Average*} \\
    miliseconds & 80 & 160 & 320 & 400 & 80 & 160 & 320 & 400 & 80 & 160 & 320 & 400 & 80 & 160 & 320 & 400 & 80 & 160 & 320 & 400\\ 
    \hline
    
    Zero-velocity & 0.39 & 0.68 & 0.99 & 1.15 
& \underline{0.27} & 0.48 & 0.73 & 0.86
& {\bf 0.26} & \underline{0.48} & \underline{0.97}  & \underline{0.95}
& \underline{0.31} & 0.67 & \underline{0.94} & \underline{1.04}
& 0.40 & 0.71 & 1.07 & 1.21\\




Tang et al. \cite{tang2018long} & \underline{0.32} & 0.53 & \underline{0.69} & \underline{0.77} 
& - & - & - & -
& - & - & - & -
& \underline{0.31} & \underline{0.66} & 0.97 & \underline{1.04}
& \underline{0.39} & \underline{0.68} & \underline{1.01} & \underline{1.13}\\
\hline 

Ours-ADD ($T=40$)  & 0.37 & \underline{0.51} & 0.77 & 0.90 
& 0.32 & \underline{0.44} & \underline{0.70} & \underline{0.82} 
& 0.36 & 0.54 & 1.02 & 0.96 
& 0.40 & 0.72 & 1.09 & 1.21
& 0.50 & 0.74 & 1.09 & 1.21\\





Ours-FN ($T=40$)  & {\bf 0.21} & {\bf 0.33} & {\bf 0.54} & {\bf 0.61} 
& {\bf 0.20} & {\bf 0.31} & {\bf 0.53} & {\bf 0.67} 
& \underline{0.28} & {\bf 0.47} & {\bf 0.83} & {\bf 0.86} 
& {\bf 0.30} & {\bf 0.61} & {\bf 0.84} & {\bf 0.94}
& {\bf 0.35} & {\bf 0.59} & {\bf 0.92} & {\bf 1.06}\\

\bottomrule
    
    \end{tabular}
    }
}
\\
\subfloat[Short-term prediction with 50 input frames and normalized angles by the standard deviation.]
{
    \resizebox{\textwidth}{!}{
        
\begin{tabular}{l|c c c c|c c c c|c c c c|c c c c||c c c c}
    \toprule
    & \multicolumn{4}{c|}{Walking} 
    & \multicolumn{4}{c|}{Eating} 
    & \multicolumn{4}{c|}{Smoking} 
    & \multicolumn{4}{c||}{Discussion} 
    & \multicolumn{4}{c}{Average*} \\
    miliseconds & 80 & 160 & 320 & 400 & 80 & 160 & 320 & 400 & 80 & 160 & 320 & 400 & 80 & 160 & 320 & 400 & 80 & 160 & 320 & 400\\ 
    \hline
  
Zero-velocity & 0.39 & 0.68 & 0.99 & 1.15 
& 0.27 & 0.48 & 0.73 & 0.86
& {\bf 0.26} & \underline{0.48} & 0.97  & 0.95 
& 0.31  & 0.67  & 0.94  & 1.04
& 0.40 & 0.71 & 1.07 & 1.21\\

 Res-Seq2Seq (sup.) \cite{martinez2017human} & \underline{0.28} & 0.49 & 0.72 & 0.81 
 & 0.23  & 0.39 & 0.62 & 0.76
 & 0.33 & 0.61 & 1.05 & 1.15 
 & 0.31  & 0.68 & 1.01 & 1.09 
 & 0.36  & 0.67  & 1.02  & 1.15 \\
 
VGRU-rl \cite{gopalakrishnan2018neural} & 0.34 & 0.47 & 0.64  & 0.72  
 & 0.27 & 0.40 & 0.64 & 0.79 
 & 0.36 & 0.61 & \underline{0.85}  & 0.92
 & 0.46 & 0.82 & 0.95  & 1.21 
 & - & - & - & -\\
 
AGED (w/o adv) \cite{gui2018adversarial} & \underline{0.28} & \underline{0.42} & 0.66 & 0.73 
& 0.22  & 0.35 & 0.61  & 0.74  
& 0.30  & 0.55  & 0.98 & 0.99 
& \underline{0.30} & 0.63  & 0.97 & 1.06 
& \underline{0.32}  & \underline{0.62} & 0.96 & 1.07 \\

AGED \cite{gui2018adversarial} 
& {\bf 0.22} & {\bf 0.36} & {\bf 0.55} & \underline{0.67} 
& {\bf 0.17} & {\bf 0.28} & {\bf 0.51} & {\bf 0.64} 
& \underline{0.27} & {\bf 0.43} & {\bf 0.82} & \underline{0.84} 
& {\bf 0.27}  & {\bf 0.56} &  {\bf 0.76} &  {\bf 0.83}
& {\bf 0.31} & {\bf 0.54} & {\bf 0.85} & {\bf 0.97} \\



\hline 
Ours-ADD ($T=60$)& 
0.30 & 0.45 & 0.74 & 0.88 
& \underline{0.21} & 0.37 & 0.65 & 0.78 
& 0.31 & 0.50 & 0.95 & 0.89
& 0.33 & 0.65 & 0.91 & 1.03
& 0.38 & 0.64 & 0.99 & 1.12\\


Ours-FN ($T=60$)  & 
0.29 & {\bf 0.36}  & \underline{0.57} & {\bf 0.64} 
& 0.24 & \underline{0.32}  & \underline{0.52} & \underline{0.67} 
& 0.36 & 0.51 & \underline{0.85}  & {\bf 0.83}
& 0.33 & \underline{0.60}  & \underline{0.84} & \underline{0.95}  
& 0.41 & \underline{0.62} & \underline{0.92} & \underline{1.03}\\


\bottomrule
    
    \end{tabular}
    }\label{Table:short-term-50-input}
}\\
\end{table*}
\egroup

We trained our method on the Human3.6M dataset \cite{ionescu2014human3} for each of the following three tasks: (1) short-term motion prediction, (2) long-term motion prediction and motion generation, and (3) action classification and label recovery.
For each, we perform pattern completion with both a forward neural network with a single dense layer (FN) and vector addition (ADD).
We distinguish motion generation from long-term motion prediction by requiring a generative model to be able to output multiple different valid results. We measure the performance of all short-term prediction methods using the community standard metric: mean joint angle error. 

\subsection{Baselines}       \label{sec:Baselines}

We follow the same evaluation method for short-term prediction as in \cite{fragkiadaki2015recurrent, jain2016structural, martinez2017human, tang2018long, li2018convolutional, pavllo2018quaternet, gopalakrishnan2018neural, gui2018adversarial}. We cite the results from the most relevant 
works to compare with our method, which are Res-Seq2Seq \cite{martinez2017human}, the model by Tang et al. \cite{tang2018long}, VGRU-rl \cite{gopalakrishnan2018neural} and AGED \cite{gui2018adversarial} which is the current state of the art. We also compare against the naive zero-velocity baseline proposed by \cite{martinez2017human} and use their code to generate long-term predictions. 

\subsection{Data preprocessing}           \label{sec:Dataset}

Following the same settings as our baselines,
we down-sample the dataset from 50 to 25 frames per second (fps), and use subject 5 for testing and the rest for training. Joint angles with small standard deviation are ignored, resulting in an input size of 54. 

We use two normalization methods depending on the baseline that we are comparing against: (1) subtract the mean and normalized between -1 and 1, which is used in \cite{tang2018long}, and (2) subtract the mean and divide by the standard deviation, which is used by the other methods \cite{martinez2017human, gopalakrishnan2018neural, gui2018adversarial}.

\subsection{Training Procedure}  \label{sec:Training}

For the representation learning step, we use gated recurrent unit (GRU) for our H-Seq2SeqsAE. For the higher-level encoder, we use $tanh$ activation. For the rest, we use $tanh$ activation when the data is normalized between -1 and 1, otherwise, we use $linear$ activation. We train using Nadam optimizer with a learning rate of $8\mathrm{e}{-4}$ and a decay rate of   $4\mathrm{e}{-3}$. We use a batch size of 64, 5-fold cross-validation and $1\mathrm{e}{4}$ samples per epoch, for 300 epochs.

\begin{figure}[t]       
\centering          
  \includegraphics[width=0.9\linewidth]{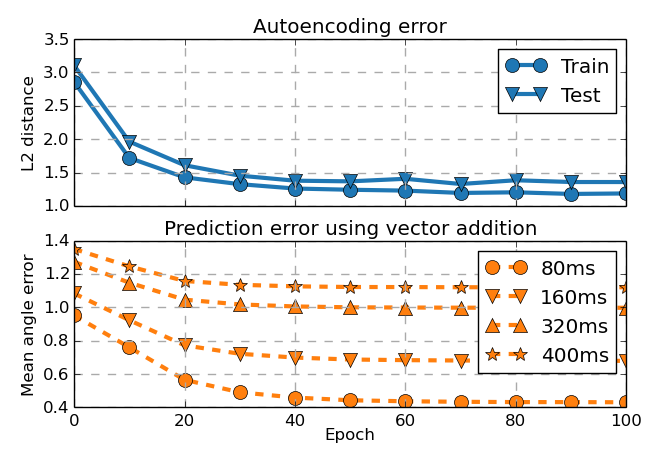}
  \caption{Convergence of autoencoding error and short-term prediction error is observed when training H-Seq2SeqsAE with $T=60$ and $\tau = 10$. Vector addition is used as the pattern completion function for this prediction task, and it is computed using 1000 training samples. Note that the autoencoding error includes the global rotations and transitions while the prediction error does not.} \label{fig:training}
\end{figure}

Except Tang et al. \cite{tang2018long} which uses 30 input time-steps and no label information, all compared methods have reported results for short-term prediction using 50 input time-steps, 10 output time-steps and appended action label information. Hence, we train two H-Seq2Seqs-AE, one with $T=40$ and 1024 latent dimensions, the other with $T=60$, 1500 latent dimensions and label information. We use $\tau = 10$ for both. Fig. \ref{fig:training} shows an example of a training curve for the latter. The latter model is also used for long-term prediction and motion generation.

For the pattern completion step with a single dense layer network, we train such network to map encodings of sequences of length 30 or 50 to their corresponding encodings of sequences of length 40 or 60, respectively, for short-term prediction. For long-term prediction and motion generation, we train with 10 input time-steps and 50 output time-steps. We use the same settings as for training H-Seq2SeqsAE, except a faster step-decay with a rate of 0.5 and 50 epochs. 

For action classification, we train a separate H-Seq2Seqs-AE with $T=40$, $\tau=10$, 1024 latent dimensions and label information with only walking and sitting actions. Our pattern completion function maps unlabeled to labeled motions of length 40.

\subsection{Short-term motion prediction}          \label{sec:Result}
Table \ref{Table:result40} shows our results for short-term prediction compared against baseline methods. We observe that our method is better than all approaches that utilize stationary loss functions. This includes the core approach of the AGED~\cite{gui2018adversarial} method. However, the unique addition of adversarial loss within that method has led to boosted performance -- a feature we have not yet implemented in our method, but which could easily augment the core improvements demonstrated here.

\bgroup
\def\tabcolsep{3pt}
\begin{table}
\centering
\caption{Short-term motion prediction for some notoriously challenging aperiodic actions where our method outperforms all baselines.} \label{Table:all40}
\subfloat[With 30 input frames and normalized angles between -1 and 1.]
{
    \resizebox{\linewidth}{!}{
        \begin{tabular}{l|c c c c|c c c c|c c c c}
\toprule
& \multicolumn{4}{c|}{Greeting} 
& \multicolumn{4}{c|}{Sitting} 
& \multicolumn{4}{c}{Taking Photo} 
\\

miliseconds & 80 & 160 & 320 & 400 
& 80 & 160 & 320 & 400 
& 80 & 160 & 320 & 400 
\\
\hline

Zero-velocity 
& \underline{0.54}  & 0.89 & 1.30 & 1.49
& {\bf 0.40} & \underline{0.63} & 1.02 & 1.18
& \underline{0.25}  & 0.51 & 0.79 & 0.92
\\

Tang et al. \cite{tang2018long}
& \underline{0.54}  & 0.87 & 1.27 & 1.45
& - & - & - & -
& 0.27 & 0.54 & 0.84 & 0.96
\\
\hline




Ours-ADD
& 0.57 & \underline{0.85}  & \underline{1.26}  & \underline{1.44}
& \underline{0.49}  & 0.67 & \underline{1.01} & \underline{1.16}
& 0.29 & \underline{0.49}  & \underline{0.74} & \underline{0.87}\\


Ours-FN
& {\bf 0.40}  & {\bf 0.69} & {\bf 1.11} & {\bf 1.28}  
& {\bf 0.40} & {\bf 0.58}  & {\bf 0.90} & {\bf 1.09} 
& {\bf 0.24}  & {\bf 0.45}  & {\bf 0.67}  & {\bf 0.77}  
\\

\bottomrule
    
    \end{tabular}

    }
}\\
\subfloat[With 50 input frames and normalized angles by the standard deviation.]
{
    \resizebox{\linewidth}{!}{
        \begin{tabular}{l|c c c c|c c c c|c c c c}
\toprule
& \multicolumn{4}{c|}{Greeting} 
& \multicolumn{4}{c|}{Sitting} 
& \multicolumn{4}{c}{Taking Photo} 
\\

miliseconds & 80 & 160 & 320 & 400 
& 80 & 160 & 320 & 400 
& 80 & 160 & 320 & 400 
\\
\hline

Zero-velocity 
& 0.54  & 0.89 & 1.30  & 1.49
& \underline{0.40} & 0.63 & 1.02 & \underline{1.18}
& 0.25 & 0.51  & \underline{0.79} & 0.92 
\\

Res. \cite{martinez2017human} & 0.75 & 1.17 & 1.74 & 1.83 &
0.41 & 1.05 & 1.49 & 1.63 & 
\underline{0.24} & 0.51 & 0.90 & 1.05

\\

AGED (w/o adv.) \cite{gui2018adversarial} & 0.61 & 0.95 & 1.44 & 1.61
& 0.46 & 0.87 & 1.23 & 1.51
& \underline{0.24} & 0.52 & 0.92 & 1.01 \\

AGED \cite{gui2018adversarial} & 0.56 & 0.81  & 1.30  & 1.46  & 
0.41 & 0.76 & 1.05 & 1.19 & 
{\bf 0.23} & \underline{0.48}  & 0.81 & 0.95 
\\

\hline

Ours-ADD
& \underline{0.47} & \underline{0.78} & \underline{1.21} & \underline{1.40}
& {\bf 0.37} & {\bf 0.58} & {\bf 0.94} & {\bf 1.10}
& {\bf 0.23} & {\bf 0.46} & {\bf 0.69} & \underline{0.80}\\



Ours-FN
& {\bf 0.46} & {\bf 0.74} & {\bf 1.14} & {\bf 1.34} 
& 0.43 &  \underline{0.62} & {\bf 0.94} & {\bf 1.10} 
& 0.31 & 0.49 & {\bf 0.69} & {\bf 0.79}\\


\bottomrule
    
    \end{tabular}

    }\label{Table:all40-b}
}\\
\end{table}
\egroup

Our method sometimes has difficulty in the first few output time-steps. This is expected since we are adding the residual angle to reconstructed poses rather than the last pose from the original input sequence as done in \cite{martinez2017human, gui2018adversarial}. However, our method excels for longer temporal horizons. We also observe that ADD outperforms FN on the first few predicted frames when the data is normalized by the standard deviation. 

One advantage of our model is that it can capture the structure in several aperiodic motions better than our baselines. 
Prior works have difficulty in modeling complicated and highly stochastic motions that even the zero-velocity baseline can easily outperform them, as observed by Martinez et al. \cite{martinez2017human}. The adversarial loss in AGED ~\cite{gui2018adversarial} also leads to a significant performance improvement in aperiodic motions, but here we outperform them all, as can be seen in the tasks greeting, sitting and taking photo (see Table \ref{Table:all40}).

\subsection{Long-term motion prediction and generation}             \label{sec:MG}

\begin{figure*}[h]  
\centering
\subfloat[Greeting.]
{
    \includegraphics[height=2.35in]{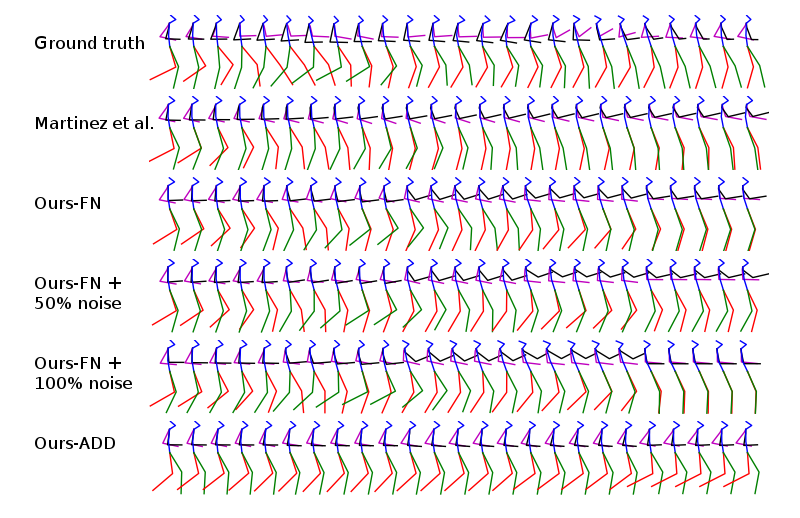}
    \label{fig:greeting_long_term}
}
\subfloat[Walking.]
{
    \includegraphics[height=2.35in]{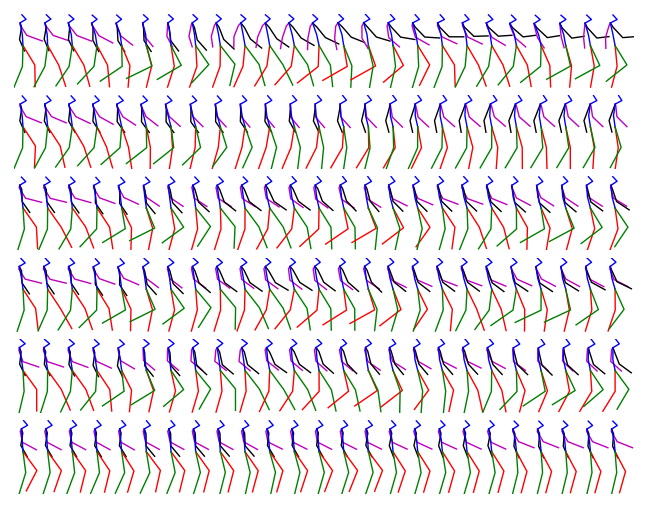}
    \label{fig:walking_long_term}
}
    \caption{Long-term motion prediction and generation of 50 time-steps for greeting and walking. Note that we skip every second frame.}
    \label{fig:generation}
\end{figure*}

Fig. \ref{fig:generation} shows our qualitative results for 
long-term motion prediction and motion generation with outputs of 50 time-steps compared with the long-term predictions by \cite{martinez2017human}. For these tasks, we use the same model that resulted Table \ref{Table:short-term-50-input}, and the input sequence length is set to 10 time-steps rather than 50 time-steps. Martinez et al. \cite{martinez2017human} propose the hypothesis that using MSE as the loss function forces the prediction to converge to a mean pose. Although our MAE loss has similar theoretical properties to MSE, we observe that our method can produce more plausible motions over a longer time horizon, even for aperiodic actions like greeting. We also observe that ADD often generates motionless sequences, albeit it can preserve some general structure of the motions. 

To obtain diverse generated solutions, we demonstrate that we can generate different motion sequences by adding noise to the output of our forward network. The amount of noise is computed using the standard deviation of the distance $d$ (see Section \ref{sec:RepMatching-Predict}). This results in a slight variation for our long-term prediction (see Fig. \ref{fig:generation}), but the latter is less smooth and may contain unnatural poses. 

\begin{figure}[t]       
\centering          
  \includegraphics[width=0.8\linewidth]{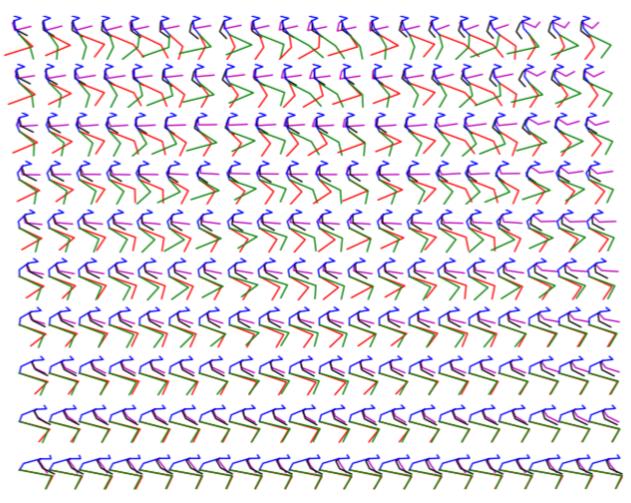}
  \caption{Interpolation between walking (top row) and sitting (bottom row). Each row represents a generated motion sequence using our model. Each column appears to be a smooth and realistic motion of a person sitting down.} \label{fig:interpolation}
\end{figure}

Our method can also demonstrate motion generation using interpolation on the latent representation space. In Fig. \ref{fig:interpolation}, given a walking and a sitting motion, we generate 8 motion sequences between the two, which can be combined to create smooth and realistic motion of a person sitting down. 

Animations and quantitative evaluations of our results are available on our project webpage\footnote{\url{http://www.cim.mcgill.ca/~yxu219/human-motion-prediction.html}}.

\subsection{Action classification and label recovery}         \label{sec:ActionClass}

To our knowledge, we are the first to perform action classification using solely the Human3.6M dataset. Prior works on skeleton-based action classification either use other datasets entirely \cite{butepage2017deep, presti20163d} or combine Human3.6M with additional data \cite{luvizon20182d} for training.
Human3.6M is a difficult dataset for action classification due to the large variety of poses and motions that overlap between the action categories. 
Therefore, we choose to select only two actions to perform action classification and label recovery on: walking and sitting. 

Table \ref{Table:Classification} shows our results compared with three simple baseline methods. Our results show that our method demonstrates a high performance improvement compared to our baselines. We also observe that our performance in label recovery is slightly lower. This reflects the difference in performance between seeing and not seeing the motions without labels during training.

\subsection{Ablation studies}             \label{sec:ablation}

The core idea in our method is that the latent representation of the input can be completed to get a solution to a task.
In Table \ref{Table:ablation}, we compare pattern completion and pattern matching for short-term prediction. Note that for pattern matching, the distance measure from Section \ref{sec:RepMatching-Predict} becomes
\begin{equation}
    d'_j(X,Y) = E(Y) - E(X).
\end{equation}
We observe that the performance boosts when applying pattern completion for both ADD and FN. The average standard deviation of $d_j$ (used in pattern completion) is also smaller than $d'_j$. Our results also show that our pattern completion approach outperforms a standard hierarchical sequence-to-sequence (H-Seq2Seq) model with encoder and decoder similar to the ones in our H-Seq2SeqsAE, while our autoencoder combined with pattern matching cannot. This demonstrates the advantage of pattern completion over pattern matching in our method.

Furthermore, Table \ref{Table:ablation} also compares our H-Seq2Seqs-AE with a basic Hierarchical Seq2Seq autoencoder (Basic) that pads input sequences with the last input frame to input varying length sequences. Our results suggest that our autoencoder results more effective latent representation for pattern completion, since the basic autoencoder cannot distinguish between long sequences with no motion towards the end and their shorter counterparts.

\bgroup
\def\tabcolsep{5pt}
\begin{table}   
\caption{Comparison of the average predicted probability between our method and our baselines for classification of walking and sitting. Note that our method outputs a valid probability vector for two classes. Our baselines are Last+Dense: a single dense layer network trained to classify based on the last pose of the input motion, Flatten+Dense: a single dense layer network trained to classify based on all poses of the input motion, and GRU+Dense: a GRU unit connected with a dense layer.}   \label{Table:Classification}
\begin{center}
    \resizebox{0.9\linewidth}{!}{
        \begin{tabular}{l l|c|c}
\toprule
& & Walking & Sitting \\

\hline
 & Last+Dense & 0.61 & 0.61\\
& Flatten+Dense & 0.61 & 0.61\\
& GRU+Dense & 0.61 & 0.61 \\
\hline

\multirow{2}{*}{Label recovery} & Ours-ADD & 0.55 & 0.52 \\
& Ours-FN & {\bf 0.72} & {\bf 0.71} \\
\hline
\multirow{2}{*}{Action classification} 
& Ours-ADD & 0.63 &  0.69 \\
& Ours-FN & {\bf 0.73} & {\bf 0.74} \\
\bottomrule
\end{tabular}












    }
\end{center}
\end{table}
\egroup

\bgroup
\def\tabcolsep{3pt}
\begin{table}
\centering
\caption{Ablation analysis on the performance difference between pattern completion and pattern matching, and our H-Seq2Seqs-AE and a basic Seq2Seq autoencoder (Basic). $X_{30}$ denotes an input sequence of length 30, and $Y_{10}$, an output sequence of length 10. }
\label{Table:ablation}
\subfloat[With $T=40$ and normalized angles between -1 and 1.]
{
    \resizebox{\linewidth}{!}{
        \begin{tabular}{l|l|c c c c|c | c}
\toprule

\multicolumn{2}{c|}{}
& \multicolumn{4}{c|}{Average}
& \multirow{2}{*}{Mean STD}  
& \multirow{2}{*}{STD STD} 
\\
\multicolumn{2}{r|}{miliseconds} 
& 80 & 160 & 320 & 400 
& &
\\
\hline

$ X_{30} \longrightarrow Y_{10}$& \multirow{2}{*}{H-Seq2Seq}
& 0.47 & 0.68 & 0.95 & \underline{1.06}
&\multirow{2}{*}{-}
&\multirow{2}{*}{-}
\\

$ X_{30} \longrightarrow X_{30}Y_{10}$ &
& 0.57 & 0.73 & 0.97 & 1.07
&  & 
\\
\hline

\multirow{2}{*}{$X_{30} \longrightarrow X_{30}Y_{10}$ } 
& Basic-ADD 
& 0.39 & 0.66 & 0.98 & 1.11
&\multirow{2}{*}{0.003}
&\multirow{2}{*}{0.003}
\\
& Basic-FN 
& \underline{0.38} & \underline{0.62} & \underline{0.93} & {\bf 1.04}
& &
\\

\hline

\multirow{2}{*}{$X_{40} \longrightarrow Y_{10}$} 
& Ours-ADD 
& 1.90 & 1.84 & 1.76 & 1.72
&\multirow{2}{*}{0.017}
&\multirow{2}{*}{0.006}
\\
& Ours-FN 
& 0.59 & 0.78 & 1.08 & 1.18
& &
\\
\hline

\multirow{2}{*}{$X_{30} \longrightarrow Y_{10}$} 
& Ours-ADD 
& 1.75 & 1.72 & 1.59 & 1.50
&\multirow{2}{*}{0.015}
&\multirow{2}{*}{0.005}
\\
& Ours-FN 
& 0.41 & 0.65 & 0.99 & 1.11
& &
\\
\hline

\multirow{2}{*}{$X_{30} \longrightarrow X_{30}Y_{10}$ } 
& Ours-ADD 
& 0.50 & 0.74 & 1.09 & 1.21
&\multirow{2}{*}{0.003}
&\multirow{2}{*}{0.006}
\\
& Ours-FN 
& {\bf 0.35} & {\bf 0.59} & {\bf 0.92} & \underline{1.06}
& &
\\

\bottomrule
\end{tabular}




    }
}\\
\subfloat[With $T=60$ and normalized angles by the standard deviation.]
{
    \resizebox{\linewidth}{!}{
        \begin{tabular}{l|l|c c c c|c | c}
\toprule

\multicolumn{2}{c|}{}
& \multicolumn{4}{c|}{Average}
& \multirow{2}{*}{Mean STD}  
& \multirow{2}{*}{STD STD} 
\\
\multicolumn{2}{r|}{miliseconds} 
& 80 & 160 & 320 & 400 
& &
\\
\hline

$ X_{50} \longrightarrow Y_{10}$& \multirow{2}{*}{H-Seq2Seq}
& 0.51 & 0.70 & 0.98 & 1.09 
&\multirow{2}{*}{-}
&\multirow{2}{*}{-}
\\
$ X_{50} \longrightarrow X_{50}Y_{10}$ &
& 0.66 & 0.79 & 1.01 & 1.10 
&
&
\\
\hline

\multirow{2}{*}{$X_{50} \longrightarrow X_{50}Y_{10}$} 
& Basic-ADD 
& \underline{0.41} & 0.67 & 0.99 & 1.12
&\multirow{2}{*}{0.015}
&\multirow{2}{*}{0.007}
\\
& Basic-FN 
& 0.45 & 0.67 & \underline{0.95} & \underline{1.07} 
& &
\\
\hline

\multirow{2}{*}{$X_{50} \longrightarrow Y_{10}$} 
& Our-ADD 
& 1.73 & 1.79 & 1.88 & 1.91
&\multirow{2}{*}{0.027}
&\multirow{2}{*}{0.051}
\\
& Our-FN 
& 0.54 & 0.72 & 0.98 & 1.10
& &
\\
\hline

\multirow{2}{*}{$X_{50} \longrightarrow X_{50}Y_{10}$} 
& Our-ADD 
& {\bf 0.38} & \underline{0.64} & 0.99 & 1.12
&\multirow{2}{*}{0.018}
&\multirow{2}{*}{0.007}
\\
& Our-FN 
& \underline{0.41} & {\bf 0.62} &  {\bf 0.92} &  {\bf 1.03}
& &
\\
\bottomrule
\end{tabular}




    }
}\\
\end{table}
\egroup

\section{Discussion}             \label{sec:Discussion}
We have presented our Hierarchical Sequence-to-Sequences Autoencoder (H-Seq2SeqsAE), a new and generic representation learning model for time-series data of various lengths. Combined with our novel pattern completion approach, we have shown in  the context of skeleton-based human motion,  that the learned representations enable short-term and long-term motion prediction, motion generation, action classification and label recovery with high quality. In particular, our performance in short-term prediction is competitive with state of the arts and outperforms in certain aperiodic actions. Why can our model attain such performance without specialized architectures and external knowledge that other state-of-the-art methods use?


One possible explanation is: for some complicated and diverse data such as the Human3.6M dataset \cite{ionescu2014human3}, representation learning can extract important and robust features that are very suitable to the pattern completion approach.
Although the lower-dimensional latent space potentially provides less information to our forward network or to vector addition during the pattern completion step, the structure gain through the autoencoding process results in a simpler learning problem from the completion perspective: given input and ground truth pairs $\{(X_i,Y_i)\}_{i=1}^N$, robust features in $X_i$ are also present in $X_iY_i$. The pattern completion approach on the latent space captures these cues, stabilizing learning and allowing stronger connections across the implied sequence.

The impact of the hyperparameter $\tau$ can be further studied. Aside from understanding the properties of our latent representation space, future works also include improving H-Seq2SeqsAE, finding more effective representation learning models suitable for pattern completion and applications in other domains.

\bibliographystyle{IEEEtran}
\bibliography{mypaper}

\begin{thebibliography}{10}
\providecommand{\url}[1]{#1}
\csname url@samestyle\endcsname
\providecommand{\newblock}{\relax}
\providecommand{\bibinfo}[2]{#2}
\providecommand{\BIBentrySTDinterwordspacing}{\spaceskip=0pt\relax}
\providecommand{\BIBentryALTinterwordstretchfactor}{4}
\providecommand{\BIBentryALTinterwordspacing}{\spaceskip=\fontdimen2\font plus
\BIBentryALTinterwordstretchfactor\fontdimen3\font minus
  \fontdimen4\font\relax}
\providecommand{\BIBforeignlanguage}[2]{{%
\expandafter\ifx\csname l@#1\endcsname\relax
\typeout{** WARNING: IEEEtran.bst: No hyphenation pattern has been}%
\typeout{** loaded for the language `#1'. Using the pattern for}%
\typeout{** the default language instead.}%
\else
\language=\csname l@#1\endcsname
\fi
#2}}
\providecommand{\BIBdecl}{\relax}
\BIBdecl

\bibitem{ionescu2014human3}
C.~Ionescu, D.~Papava, V.~Olaru, and C.~Sminchisescu, ``Human3. 6m: Large scale
  datasets and predictive methods for 3d human sensing in natural
  environments,'' \emph{IEEE transactions on pattern analysis and machine
  intelligence}, vol.~36, no.~7, pp. 1325--1339, 2014.

\bibitem{bengio2013representation}
Y.~Bengio, A.~Courville, and P.~Vincent, ``Representation learning: A review
  and new perspectives,'' \emph{IEEE transactions on pattern analysis and
  machine intelligence}, vol.~35, no.~8, pp. 1798--1828, 2013.

\bibitem{barsalou2005situated}
L.~W. Barsalou, ``Situated conceptualization,'' in \emph{Handbook of
  categorization in cognitive science}.\hskip 1em plus 0.5em minus 0.4em\relax
  Elsevier, 2005, pp. 619--650.

\bibitem{fragkiadaki2015recurrent}
K.~Fragkiadaki, S.~Levine, P.~Felsen, and J.~Malik, ``Recurrent network models
  for human dynamics,'' in \emph{Proceedings of the IEEE International
  Conference on Computer Vision}, 2015, pp. 4346--4354.

\bibitem{jain2016structural}
A.~Jain, A.~R. Zamir, S.~Savarese, and A.~Saxena, ``Structural-rnn: Deep
  learning on spatio-temporal graphs,'' in \emph{Proceedings of the IEEE
  Conference on Computer Vision and Pattern Recognition}, 2016, pp. 5308--5317.

\bibitem{martinez2017human}
J.~Martinez, M.~J. Black, and J.~Romero, ``On human motion prediction using
  recurrent neural networks,'' in \emph{2017 IEEE Conference on Computer Vision
  and Pattern Recognition (CVPR)}.\hskip 1em plus 0.5em minus 0.4em\relax IEEE,
  2017, pp. 4674--4683.

\bibitem{pavllo2018quaternet}
D.~Pavllo, D.~Grangier, and M.~Auli, ``Quaternet: A quaternion-based recurrent
  model for human motion,'' \emph{arXiv preprint arXiv:1805.06485}, 2018.

\bibitem{gui2018adversarial}
L.-Y. Gui, Y.-X. Wang, X.~Liang, and J.~M. Moura, ``Adversarial geometry-aware
  human motion prediction,'' in \emph{ECCV}, 2018, pp. 823--842.

\bibitem{lin2018human}
X.~Lin and M.~R. Amer, ``Human motion modeling using dvgans,'' \emph{arXiv
  preprint arXiv:1804.10652}, 2018.

\bibitem{li2018convolutional}
C.~Li, Z.~Zhang, W.~S. Lee, and G.~H. Lee, ``Convolutional sequence to sequence
  model for human dynamics,'' in \emph{Proceedings of the IEEE Conference on
  Computer Vision and Pattern Recognition}, 2018, pp. 5226--5234.

\bibitem{tang2018long}
Y.~Tang, L.~Ma, W.~Liu, and W.~Zheng, ``Long-term human motion prediction by
  modeling motion context and enhancing motion dynamic,'' \emph{arXiv preprint
  arXiv:1805.02513}, 2018.

\bibitem{li2015hierarchical}
J.~Li, M.-T. Luong, and D.~Jurafsky, ``A hierarchical neural autoencoder for
  paragraphs and documents,'' \emph{arXiv preprint arXiv:1506.01057}, 2015.

\bibitem{doersch2016tutorial}
C.~Doersch, ``Tutorial on variational autoencoders,'' \emph{arXiv preprint
  arXiv:1606.05908}, 2016.

\bibitem{goodfellow2014generative}
I.~Goodfellow, J.~Pouget-Abadie, M.~Mirza, B.~Xu, D.~Warde-Farley, S.~Ozair,
  A.~Courville, and Y.~Bengio, ``Generative adversarial nets,'' in
  \emph{Advances in neural information processing systems}, 2014, pp.
  2672--2680.

\bibitem{mikolov2013efficient}
T.~Mikolov, K.~Chen, G.~Corrado, and J.~Dean, ``Efficient estimation of word
  representations in vector space,'' \emph{arXiv preprint arXiv:1301.3781},
  2013.

\bibitem{mikolov2013distributed}
T.~Mikolov, I.~Sutskever, K.~Chen, G.~S. Corrado, and J.~Dean, ``Distributed
  representations of words and phrases and their compositionality,'' in
  \emph{Advances in neural information processing systems}, 2013, pp.
  3111--3119.

\bibitem{pennington2014glove}
J.~Pennington, R.~Socher, and C.~Manning, ``Glove: Global vectors for word
  representation,'' in \emph{Proceedings of the 2014 conference on empirical
  methods in natural language processing (EMNLP)}, 2014, pp. 1532--1543.

\bibitem{holden2015learning}
D.~Holden, J.~Saito, T.~Komura, and T.~Joyce, ``Learning motion manifolds with
  convolutional autoencoders,'' in \emph{SIGGRAPH Asia 2015 Technical
  Briefs}.\hskip 1em plus 0.5em minus 0.4em\relax ACM, 2015, p.~18.

\bibitem{butepage2017deep}
J.~B{\"u}tepage, M.~J. Black, D.~Kragic, and H.~Kjellstr{\"o}m, ``Deep
  representation learning for human motion prediction and classification,'' in
  \emph{IEEE Conference on Computer Vision and Pattern Recognition
  (CVPR)}.\hskip 1em plus 0.5em minus 0.4em\relax IEEE, 2017, p. 2017.

\bibitem{sutskever2014sequence}
I.~Sutskever, O.~Vinyals, and Q.~V. Le, ``Sequence to sequence learning with
  neural networks,'' in \emph{Advances in neural information processing
  systems}, 2014, pp. 3104--3112.

\bibitem{gopalakrishnan2018neural}
A.~Gopalakrishnan, A.~Mali, D.~Kifer, C.~L. Giles, and A.~G. Ororbia, ``A
  neural temporal model for human motion prediction,'' \emph{arXiv preprint
  arXiv:1809.03036}, 2018.

\bibitem{presti20163d}
L.~L. Presti and M.~La~Cascia, ``3d skeleton-based human action classification:
  A survey,'' \emph{Pattern Recognition}, vol.~53, pp. 130--147, 2016.

\bibitem{luvizon20182d}
D.~C. Luvizon, D.~Picard, and H.~Tabia, ``2d/3d pose estimation and action
  recognition using multitask deep learning,'' in \emph{The IEEE Conference on
  Computer Vision and Pattern Recognition (CVPR)}, vol.~2, 2018.

\end{thebibliography}
%



\end{document}